\documentclass{article}

\usepackage{fullpage}
\date{}

\usepackage{amsmath,amssymb,amsthm,amsfonts}
\usepackage{graphicx}
\usepackage{booktabs}
\theoremstyle{plain}
\newtheorem{theorem}{Theorem}[section]
\theoremstyle{definition}
\newtheorem{definition}[theorem]{Definition}

\usepackage{enumerate}
\usepackage{algorithm}
\usepackage{cite,url,hyperref}

\providecommand{\keywords}[1]
{
  \small	
  \textbf{\textit{Keywords---}} #1
}

\title{An Alternative Practice of Tropical Convolution to Traditional Convolutional Neural Networks}
\author{
  Shiqing Fan \\
    School of Informatics\\
  Xiamen University\\
  Xiamen,  Fujian 361005, China \\
   \texttt{loy.fsq@gmail.com} \\
   \and
Liying Liu \\
    School of Informatics\\
  Xiamen University\\
  Xiamen,  Fujian 361005, China \\
 \texttt{llyxmu185@stu.xmu.edu.cn}
  \and
   Ye Luo\footnote{Corresponding Author. \newline This work is supported by National Natural Science Foundation of China (Grant No. 61875169).}\\
 School of Informatics\\
  Xiamen University\\
  Xiamen,  Fujian 361005, China \\
  \texttt{luoye@xmu.edu.cn} \\
}

\begin{document}
\maketitle

\begin{abstract}
Convolutional neural networks (CNNs) have been used in many machine learning fields. In practical applications, the computational cost of convolutional neural networks is often high with the deepening of the network and the growth of data volume, mostly due to a large amount of multiplication operations of floating-point numbers in convolution operations. To reduce the amount of multiplications, we propose a new type of CNNs called Tropical Convolutional Neural Networks (TCNNs) which are built on tropical convolutions in which the multiplications and additions in conventional convolutional layers are replaced by additions and min/max operations respectively. In addition, since tropical convolution operators are essentially nonlinear operators, we expect TCNNs to have higher nonlinear fitting ability than conventional CNNs. In the experiments, we test and analyze several different architectures of TCNNs for image classification tasks in comparison with similar-sized conventional CNNs. The results show that TCNN can achieve higher expressive power than ordinary convolutional layers on the MNIST and CIFAR10 image data set. In different noise environments, there are wins and losses in the robustness of TCNN and ordinary CNNs.
\newline
\\
\normalsize \textbf{CCS CONCEPTS ·} Computing methodologies \textbf{·} Machine learning \textbf{·} Machine learning approaches \textbf{·} Neural networks
\newline
\end{abstract}

\section{Introduction}\label{1}

In recent years, convolutional neural networks (CNNs) have been widely used in computer vision field of machine learning \cite{10.1561/2000000039} tasks related to images and videos, such as image classification \cite{6248110,HAN201843}, face verification \cite{NIPS2014_e5e63da7,CROSSWHITE201835}, object detection \cite{Cai_2018}, image Segmentation \cite{Chen_2016_CVPR,8270673} and so on. But with the deepening of the CNN layers, sometimes there will be billions of floating point multiplications that make the computational cost of the convolutional neural network extremely high. Although the emergence of graphics processing unit (GPU) has increased the speed of deep convolutional networks, high-end GPU cards require many other hardware support, which prevents them from being easily installed on smart watches, mobile phones and other mobile devices. Therefore, research on efficient forward reasoning deep neural network is very necessary.

To speed up CNNs and cut down energy consumption, researchers have proposed many approaches, such as changing the weight value to make the multiplication into a bool operation  \cite{NIPS2015_3e15cc11,NIPS2016_d8330f85}, or changing the traditional convolution operation to reduce the large number of multiplications in the network \cite{kim2021effects,Chen_2020_CVPR}. Among these attempts, the application of binary network can reduce computational cost, but it will also cause the network to fail to maintain the original recognition accuracy and converge slowly during training. Moreover, in directly changing the direction of the convolution operation, some papers use different methods to reduce or eliminate the multiplication in the convolution, such as changing the metric operation between the input image and the convolutional kernel to a metric where the multiplication is not dominant. Our work continues this direction, trying to find a structure as a replacement of the traditional convolution operation which can greatly reduce the number of multiplications from an algebra that has been proven to have global expression capabilities. 

In network calculation \cite{858696}, a convolution operation using minplus algebra is commonly used to estimate the performance of communication network, which replaces the nonlinear system with a linear system with the lower limit of the nonlinear system to calculate the worst-case performance index. In such a minplus convolution, the multiplication and addition operations in a conventional convolution are replaced by addition and minimum operations respectively. The arithmetic behind this convolution structure is directly related to the mathematical theory of tropical semi-rings, which is fast developing area in pure and applied mathematics \cite{litvinov2009tropical}. 

It is well-known that executing additions is usually faster than executing multiplications with less energy consumption. 
Therefore, the min-plus convolution is a theoretically feasible solution to be applied to CNNs as a replacement of conventional convolutions to reduce the overall computation cost.

In this article, we propose a new type of CNNs called Tropical Convolutional Neural Networks (TCNNs) in which the tropical convolution operators (including both minplus and maxplus operators) are used in the convolution layer. When applied to computer vision, a TCNN maximizes the use of addition to replace the floating point multiplication between the convolutional kernel and the pixels of input images when performing convolution computations, and takes the minimum value or maximum value across the entire sliding window as output. Throughout the process of feature extraction of the entire convolutional network, the use of addition also has the capability to measure the similarity of convolutional kernel and input image as in traditional CNNs. On multiple benchmark classification datasets such as MNIST and CIFAR10, experimental results show that TCNNs can reduce the proportion of deep neural network multiplication while achieving higher accuracy compared with traditional CNNs.

The organization of this article is as follows. In Section ~\ref{2}, we study related work on network calculus and multiplication substitution. In Section ~\ref{3}, we present the tropical convolution operations and the basic structure of TCNNs. In Section ~\ref{4}, we show several different network structure designs, and evaluate the performance of various benchmark data sets on the TCNNs, and makes a statistical comparison with traditional CNNs. Section ~\ref{5} is a summary of the paper.

\section{Related Work}\label{2}
\subsection{Min-plus Convolution}\label{2.1}

A lot of work has found that the superiority of an application of minplus algebra to convolution calculation. Because of the features of this kind of convolution, it has been widely used in different fields. One of the major applications is network calculation \cite{669170} which provides a theoretical framework for analyzing the performance guarantee of computer networks. Specifically, it can calculate the worst-case limits of delay and buffer requirements in the network. For example, Zubaydy et al. \cite{6566982} proposed a network calculus that can incorporate common statistical models of fading channels and obtain statistical boundaries for delays and backlogs across multiple nodes. And Tabatabaee et al. \cite{2020arXiv200308372M} used a network calculus approach and found a strict service curve for Interleaved Weighted Round-Robin.

\subsection{Network Demultiplication}\label{2.2}

Recently, many researchers have worked with efforts to reduce the cost of the underlying circuit by cutting down the multiplication in deep neural networks. One of the more typical methods to decrease multiplication is to improve network parameters. For example, Courbariaux et al. \cite{NIPS2015_3e15cc11} wrote a seminal paper introducing BinaryConnect to train the deep learning network with binary weights, in which multiply-accumulate operations are replaced by simple accumulations. Similarly, Hubara et al. \cite{NIPS2016_d8330f85} proposed BNNs, which also used binary activation besides binary weights to improve power-efficiency. On the other hand, there are some other works that directly replace the multiplication in the network calculation to lower the computational cost:  Kim et al. \cite{kim2021effects} analyzed and justified that multiplications can be approximated while additions need to be exact in CNN Multiply Accumulate operations; XNOR-nets \cite{10.1007/978-3-319-46493-0_32} used the approximate binary value in the fully connected layer to replace convolution with binary calculations; AdderNet \cite{Chen_2020_CVPR} calculated the 1-norm distance between filters and input functions to take place of multiplication. In a previous work, we proposed Min-Max-Plus Neural Networks \cite{MMP} as a new model of neural networks whose fitting ability can be more concentrated in the nonlinear part which is composed of minplus layers and maxplus layers with no multiplications performed.

\section{Methods}\label{3}
\subsection{Tropical Algebra  and Tropical Convolution Operators}\label{3.1}
Here we start with the traditional convolution, a particular kind of integral transform, usually defined as the integral of the product of the two functions ($f$ and $g$) after one is reversed and shifted as in Definition~\ref{D:conv}, where $t$ can be any domain and does not necessarily represent the time domain.

\begin{definition}\label{D:conv} 
The \emph{convolution} of two functions $f$ and $g$ is
$$
(f*g)(t) = \int_{0}^{t} f(\tau) g(t-\tau) d \tau
$$
\end{definition}

In tropical arithmetic over tropical semiring, the multiplication and addition are replace by addition and minimum (or maximum) respectively. More precisely, we have the following definition of tropical semiring. 
\begin{definition}
The \emph{tropical semiring} may refer to either the minplus algebra or the maxplus algebra (we need to use both of them in this paper). 
\begin{enumerate}[(i)]
\item The \emph{minplus algebra} is defined as
$$
\mathbb{R}_{\min}:=(\mathbb{R} \bigcup\{\infty\}, \oplus, \odot)
$$
where the \emph{tropical (lower) addition} $\oplus$ and the \emph{tropical multiplication} $\odot$ are defined respectively as $x \oplus y:=\min(x, y)$ and $x \odot y:=x+y$ for all $x,y \in \mathbb{R}_{\min}$. 
\item As a dual, the \emph{maxplus algebra} is defined as
$$
\mathbb{R}_{\max}:=(\mathbb{R} \bigcup\{-\infty\}, \boxplus, \odot)
$$
where the \emph{tropical (upper) addition} $\boxplus$ is defined  as $x \boxplus y:=\max(x, y)$  for all $x,y \in \mathbb{R}_{\max}$.
\end{enumerate}
\end{definition}

Note that $\infty$ and $-\infty$ are the identities for tropical lower and upper additions respectively. In practice, all computations are performed within $\mathbb{R}$. 

As widely used in network calculus, the minplus convolution is a notion similar to traditional convolution (Definition~\ref{D:conv}) where operations are performed in minplus algebra where multiplication and addition are replaced by tropical multiplication (addition) and tropical lower addition (minimum) respectively. 

While min-plus convolution, which is widely used in network calculus, uses plus and min operators to replace multiplication and sum operators in  respectively, and obtains the following Definition~\ref{D:min}. Similarly, we also have Max-Plus Convolution described as Definition~\ref{D:max}.

\begin{definition}\label{D:min} 
The \emph{min-plus convolution} of two functions $f$ and $g$ is defined as 
$$(f \otimes g)(t) = \inf_{0 \geq \tau \geq t} (f(\tau)+g(t-\tau)).$$
\end{definition}

In addition, there is also a notion of minplus deconvolution as the inverse operator of the convolution operation  can be expressed as Definition~\ref{D:mindeconv}.

\begin{definition}\label{D:mindeconv} 
The minplus deconvolution of two functions  $f$ and $g$ is defined as
$$(f \oslash g)(t) = \sup{\tau \geq 0} (f(t+\tau)-g(\tau))$$
\end{definition}

As a dual to minplus convolution, we may define maxplus convolution analogously. 

\begin{definition}\label{D:max} 
The maxplus convolution of two functions  $f$ and $g$ is defined as
$$(f \boxplus g)(t) = \sup_{0 \geq \tau \geq t} (f(\tau)+g(t-\tau))$$
\end{definition}

Both minplus and maxplus convolution operators can be used either individually or collaboratively in varied types of our proposed TCNNs. 

\subsection{Operators and Building Blocks}\label{3.2}

In a typical CNN (e.g., applied to a computer vision task), the convolutional layer commonly refers to the  operation of Definition~\ref{D:conv} where the domain can be higher dimensional. In particular, consider a convolutional kernel $K \in \mathbb{R}^{k \times k \times C_{in} \times C_{out}}$ in an intermediate layer of the neural network, where $C_{out}$ is the output channel size, $C_{in}$ is the input channel size and k is kernel size. The input feature is defined as $X \in \mathbb{R}^{H_{in} \times W_{in} \times C_{in}}$, where $H$ and $W$ are the height and width of the feature, respectively. And the output feature $Y \in \mathbb{R}^{H_{out} \times W_{out} \times C_{out}}$ can be computed as the convolution of $X$ and $K$ by the following equation.
\begin{equation}\label{equ:tradition}
    Y(h,w,p)= \sum_{d=0}^{C_{in}}\sum_{i,j=0}^k(X(h+i,w+j,d) \times K(i,j,d,p)) + bias
\end{equation}
where $h$, $w$, $p$ are the indices of a pixel, and $bias$ is the bias of the convolutional neural kernel.  Currently, there are many ways to change operations in convolution, but most of them contain some or many multiplication operations, which makes the computational cost in the convolutional layer high.

\begin{figure}[ht]
\centerline{\includegraphics[scale=0.2]{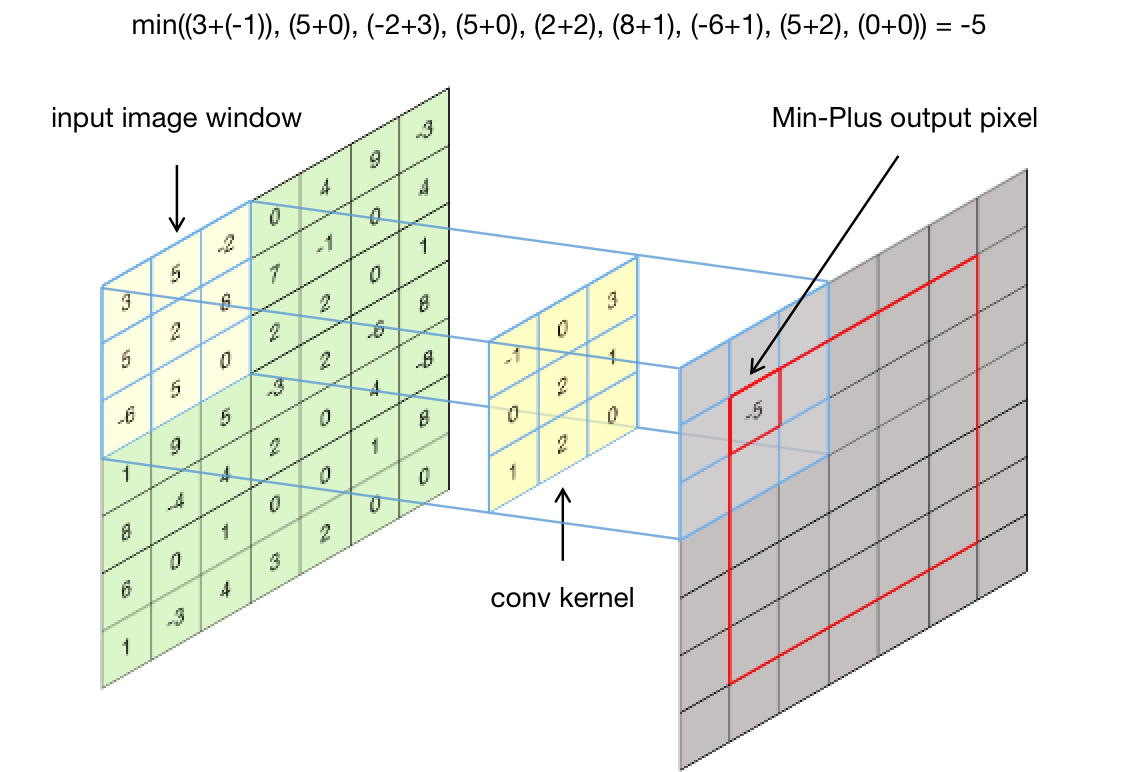} \includegraphics[scale=0.2]{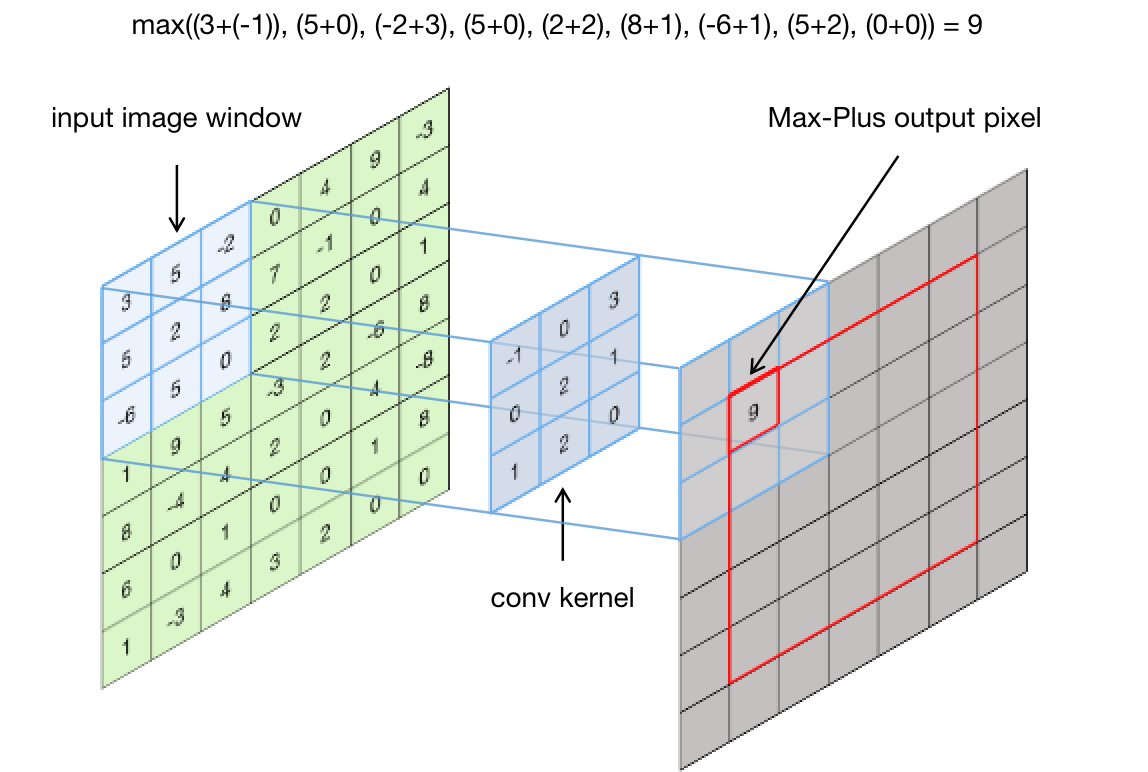}}
\vspace{.1in}
\caption{MinPlus and MaxPlus operators.}
\label{F:conv-operator}
\end{figure}

To minimize the beep of multiplication in the convolution calculation, we use the idea of minplus convolution for reference, that is, simply change the $(x \times y)$ in Equation ~\ref{equ:tradition} to $(x+y)$, which means all multiplication operations will be replaced by addition. And we also change the original process of calculating the sum in a sliding window to taking the minimum/maximum value.

The calculation process of MinPlus and MaxPlus convolution operators are shown in Figure ~\ref{F:conv-operator}. Since the two processes are very similar, we take MinPlus as an example. (For MaxPlus convolution operator, simply change the minimum to maximum.) Suppose there is an input image of size $7 \times 7$, and we set the size of the convolutional kernel to $3 \times 3$ (that is, the size of the sliding window is $3 \times 3$) and the sliding step to 1. Then the calculation process of the convolution operation in a sliding window can be divided into two steps. 
\begin{enumerate}
    \item[Step 1:] Add the $3 \times  3$ sliding window by the corresponding elements of the convolutional kernel to obtain a new $3 \times 3$ matrix, and use the minimum of the new matrix as the extracted value of this sliding window. 
    \item[Step 2:] Slide the window with a stride of 1, and after computing all the values as same as step 1, we can get the final output image of the size $5 \times 5$. 
\end{enumerate}

\subsection{The Tropical Convolution Layers}\label{3.3}

Now suppose we have a multi-channel input image. Based on the operators in Figure ~\ref{F:conv-operator}, we can get the first step of the example in Figure ~\ref{F:conv-layer}, that is, the $7\times7\times3$ input image ($3$ is $C_{in}$, the number of input channels) passes through two $3\times3$ convolutional kernels separately to obtain a $5\times5\times3$ intermediate result each. After that, we can either sum up or take the maximum value of the corresponding positions of each channel of the intermediate result to obtain the output result in one single channel. In this example, we finally get the feature map of size $5\times5\times2$, thus completing a forward process of Min(Max)Plus-Sum(Max)-Conv layer. The whole operations above can be described as the following 6 types of tropical convolutional layers:

\begin{figure}[ht]
\vspace{.1in}
\centerline{\includegraphics[scale=0.25]{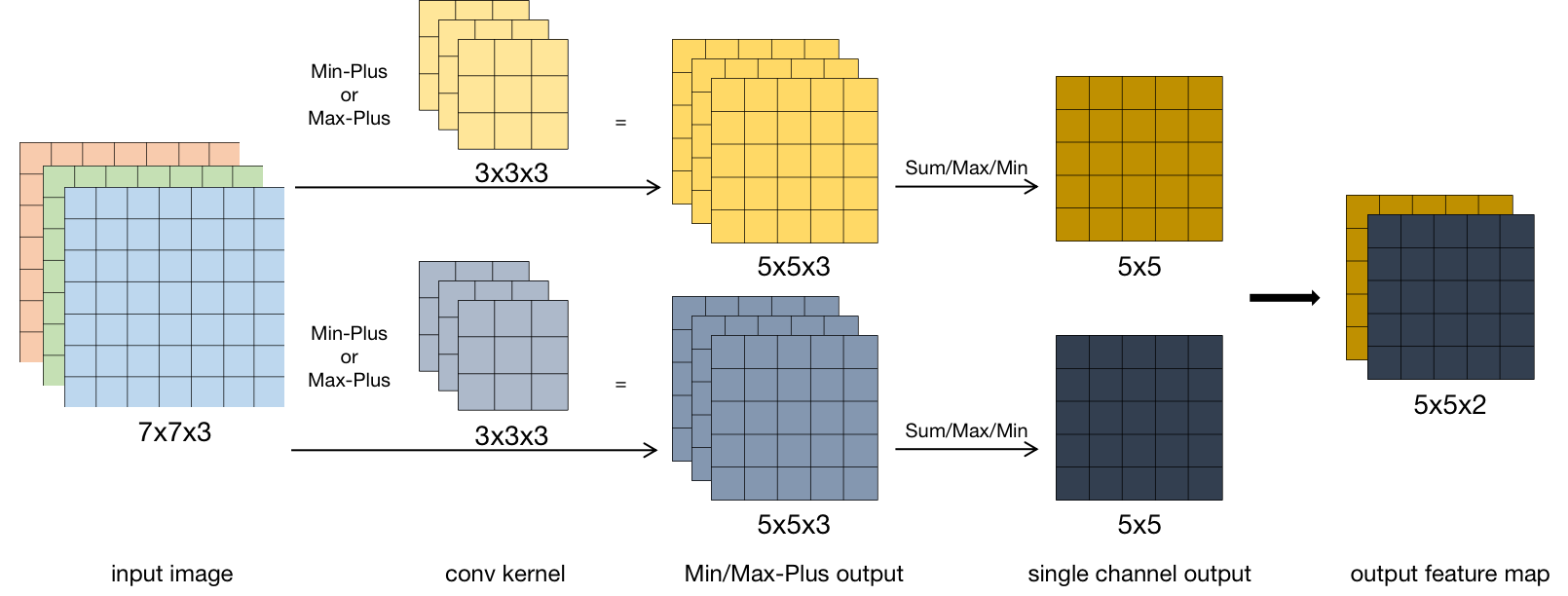}}
\vspace{.1in}
\caption{An example of tropical convolutional layer.}
\label{F:conv-layer}
\end{figure}

\begin{definition}
We define 6 types of tropical convolutional layers. 
\begin{enumerate}[(i)]
\item MinPlus-Sum-Conv Layer (MinP-S)
$$
Y(h,w,p)=\sum_{d=0}^{C_{in}}\min_{i,j=0}^k (X(h+i,w+j,d)+K(i,j,d,p))
$$

\item MaxPlus-Sum-Conv Layer (MaxP-S)
$$
Y(h,w,p)=\sum_{d=0}^{C_{in}}\max_{i,j=0}^k (X(h+i,w+j,d)+K(i,j,d,p))
$$

\item MinPlus-Max-Conv Layer (MinP-Max)
$$
Y(h,w,p)=\max_{d=0}^{C_{in}}\min_{i,j=0}^k (X(h+i,w+j,d)+K(i,j,d,p))
$$

\item MaxPlus-Max-Conv Layer (MaxP-Max)
$$
Y(h,w,p)=\max_{d=0}^{C_{in}}\max_{i,j=0}^k (X(h+i,w+j,d)+K(i,j,d,p))
$$

\item MinPlus-Min-Conv Layer (MinP-Min)
$$
Y(h,w,p)=\min_{d=0}^{C_{in}}\min_{i,j=0}^k (X(h+i,w+j,d)+K(i,j,d,p))
$$

\item MaxPlus-Max-Conv Layer (MaxP-Min)
$$
Y(h,w,p)=\min_{d=0}^{C_{in}}\max_{i,j=0}^k (X(h+i,w+j,d)+K(i,j,d,p))
$$

\end{enumerate}
\end{definition}

We note that these tropical convolution layers do not use multiplication at all. Moreover, since min/max operations are performed, unlike conventional convolution layers which are linear, tropical convolution layers are essentially nonlinear layers, and incorporating them in the network can increase the overall non-linear expression capabilities of the network.

\begin{figure}[ht]
\vspace{.1in}
\centerline{\includegraphics[scale=0.2]{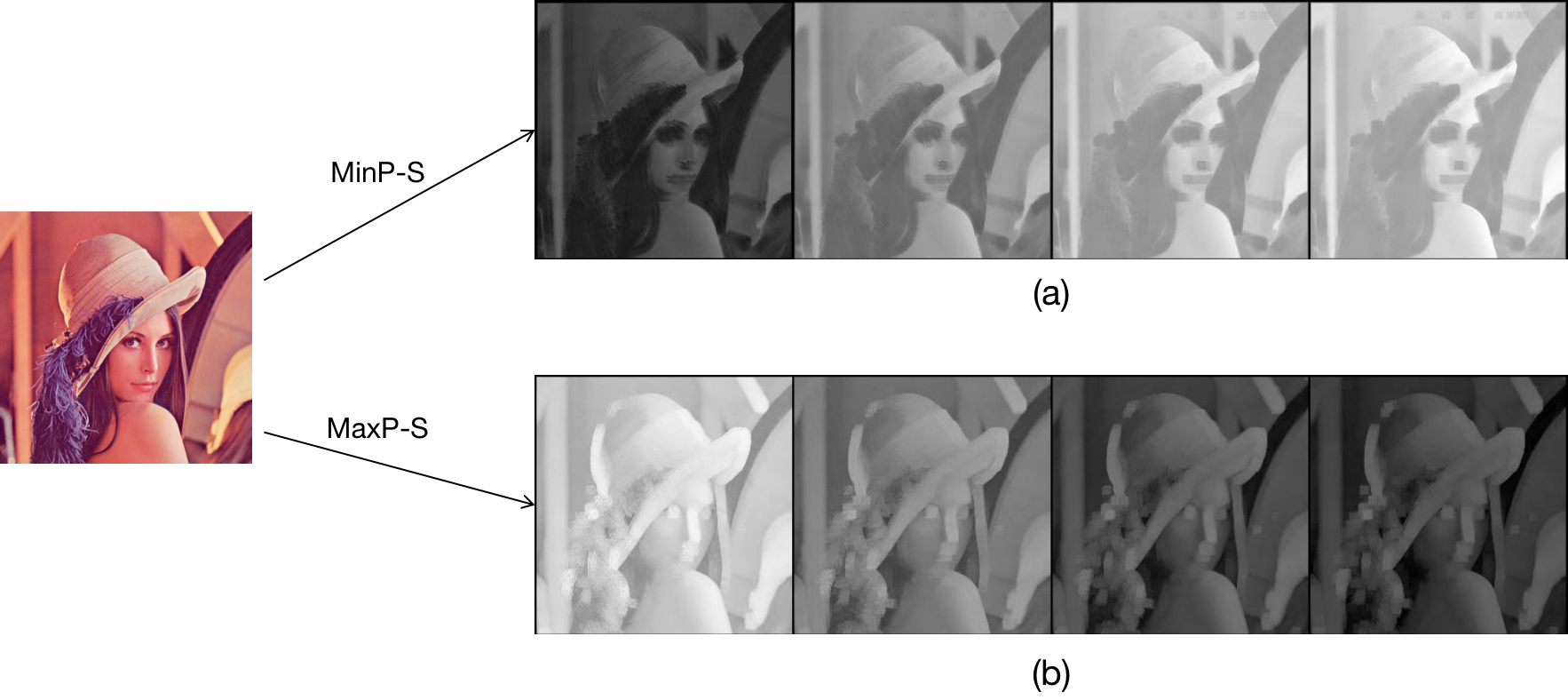}}
\vspace{.1in}
\caption{The effect of color pictures after going through the MinP-S and MaxP-S layers.}
\label{F:lena}
\end{figure}

We use a classic processing image "Lena" as an example to show the result of this color picture passing through the MinP-S and MaxP-S layers, and the results are shown in Figure~\ref{F:lena}. All the sizes of convolutional kernels in MinP-S and MaxP-S are set to $16 \times 16$, and the entry values of the kernels in four output channels are random chosen from the interval $[-1,1]$ multiplied by $1$, $0.5$, $0.2$, $0.1$ respectively. It can be seen that the result of MinP-S, shown in Figure~\ref{F:lena}.(a), and the result of MaxP-S, shown in Figure~\ref{F:lena}.(b), can extract features at different levels of the image without using multiplication. Furthermore, we observe that the darker parts of the image are displayed more prominently when the values in the convolution kernels become smaller through MinP-S, while MaxP-S shows the opposite effect that the lighter parts are prominently displayed.

\section{SOME EXPERIMENTS}\label{4}

\subsection{TCNNs}\label{4.1}

In Section~\ref{3.2}, we've defined several types of tropical convolution layers with different calculation methods not involving multiplication. In practice, we can make a variety of combinations of these convolution layers together with some linear layers and choose from them an optimized one for a specific application. Here are some of our solutions.

\begin{figure}[ht]
\vspace{.1in}
\centerline{\includegraphics[scale=0.35]{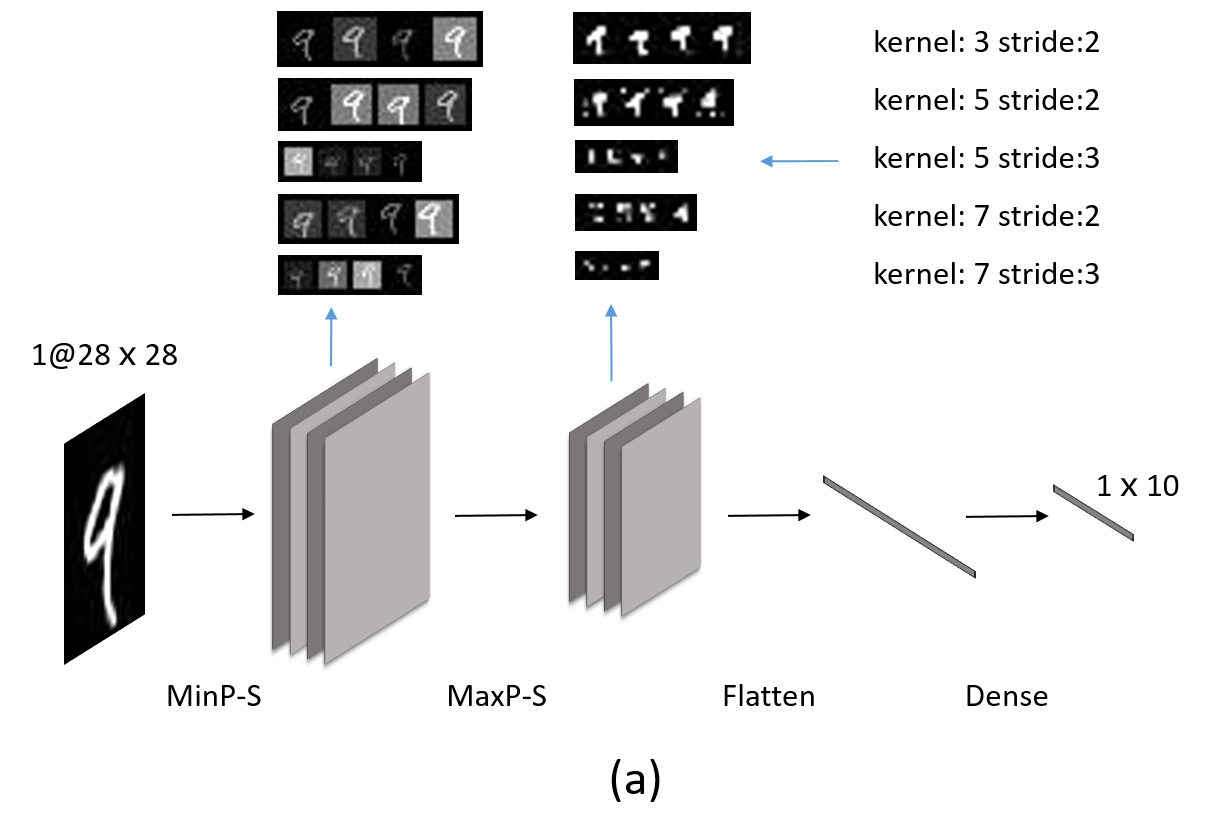}\includegraphics[scale=0.35]{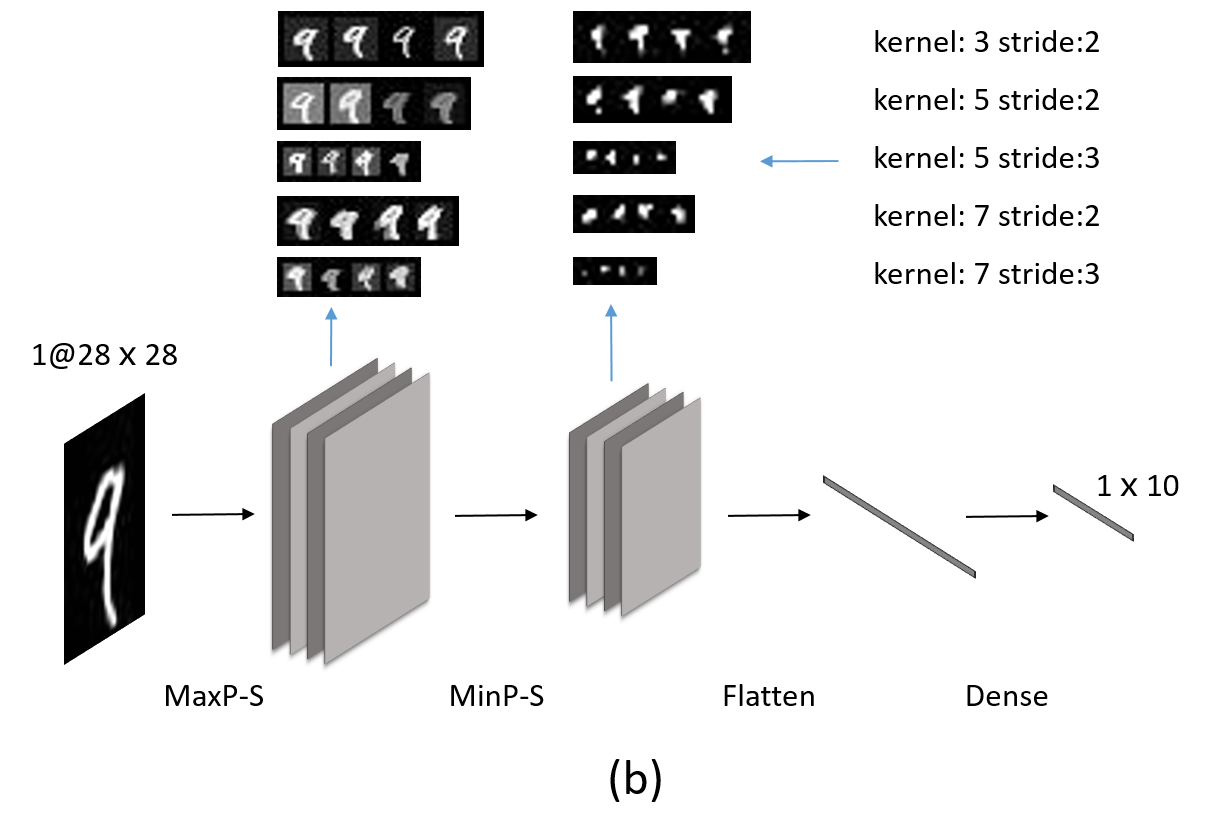}}
\vspace{.1in}
\centerline{\includegraphics[scale=0.35]{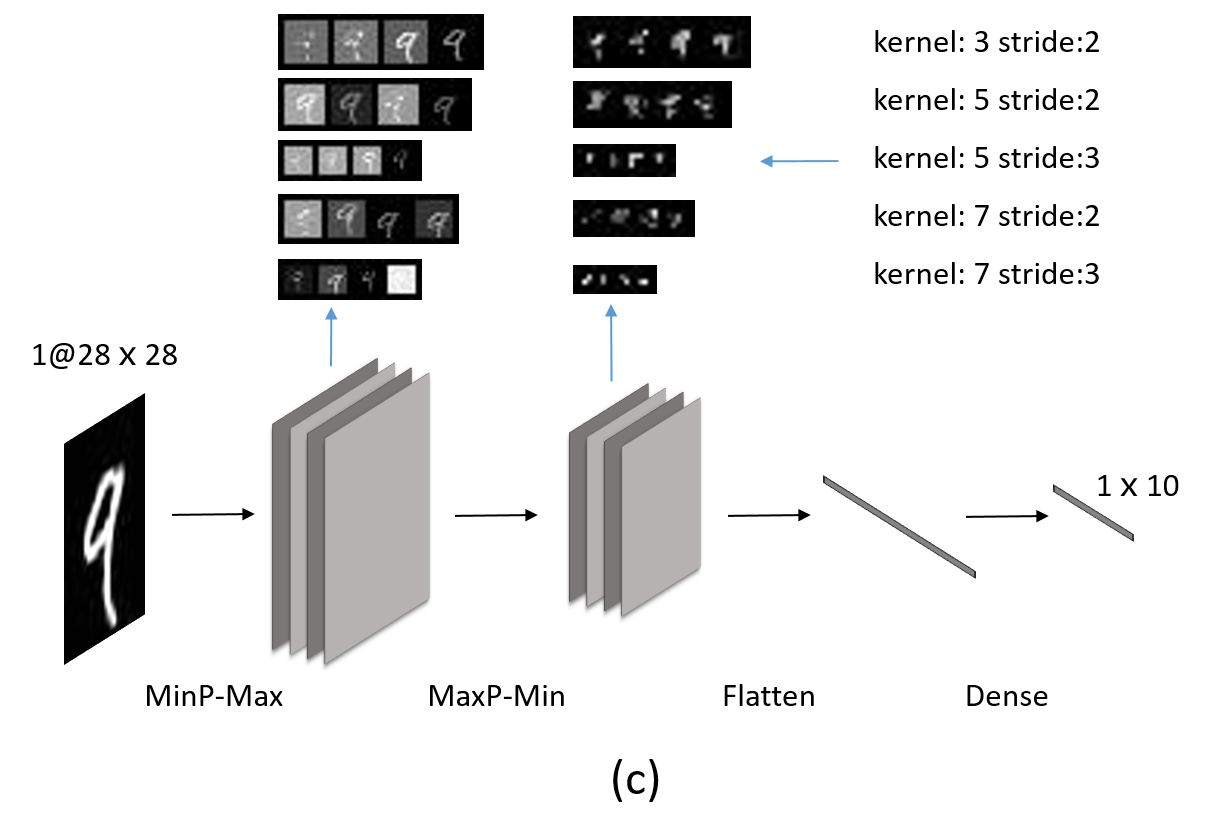}\includegraphics[scale=0.35]{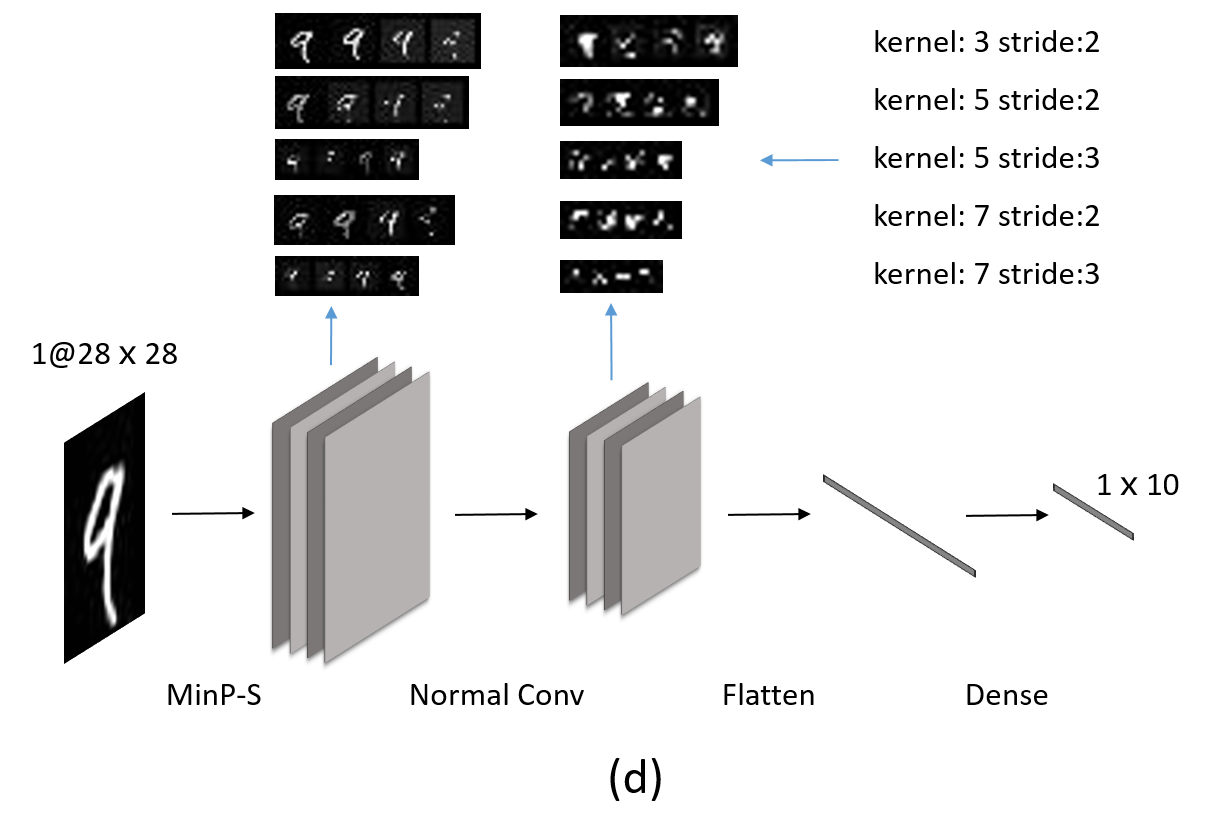}}
\caption{TCNNs.}
\label{F:nn1}
\end{figure}

Figure~\ref{F:nn1} shows four different network structures, which are composed of tropical convolution operators, linear layers, common convolution layers and so on. We use these structures to conduct preliminary experiments on handwritten digit dataset MNIST. We compare  a number of different convolutional kernel size and sliding step size combinations using visualization of the hidden layer's feature map. We observe that in the structure of Figure~\ref{F:nn1}(a) and Figure~\ref{F:nn1}(b), the first layer of MinP-S or MaxP-S provides images like different contrasts and different styles on the premise of retaining the main information of the original data. We think that this can be regarded as a data feature extraction with data enhancement function. The structure of tropical convolution makes the convolution layers have good nonlinear expression ability without activation function. We also combine tropical convolution layer with ordinary convolution layer, expecting to take advantage of the features of both tropical convolution which can provide data enhancement for the network at the first layer and  ordinary convolution which has the powerful information extraction ability for feature extraction. This structure is shown in Figure~\ref{F:nn1}(d). We will make statistics on the specific task implementation effect in the next session.

\subsection{Classification}\label{4.2}

In the final image classification results, we've also tested TCNNs with an comparison to ordinary CNN. We uniformly use a 4-layer network, the latter two layers are fixed as flatten layer and Dense layer, the first two layers use different tropical convolutional modules, the output of the first layer is unified to 15x15 4-channel feature map, and the output of the second layer is unified to $7 \times 7$. The four-channel feature map ensures that the parameters of the four networks are consistent. Among them, the Conv-Conv network only contains ordinary convolution layers and both layers have a large number of multiplications; the first two layers of (MinP-S)+(MaxP-S) and (MinP-Max)+(MaxP-Min) networks do not contain any multiplications; in (MinP-S)+(Conv) network, a tropical convolution layer  is followed by  a layer of ordinary convolution, and thus there is no multiplication in the first layer.

\begin{table}[H]
    \caption{Classification accuracy of several TCNNs on image classification tasks.}
    \label{tab:classification}
    \centerline{
    \begin{tabular}{ccl}
        \toprule
        Network Structure&MNIST&CIFAR10\\
        \midrule
        Conv+Conv+Flatten+Dense & 92.18$\%$ & 39.34$\%$ \\
        (MinP-S)+(MaxP-S)+Flatten+Dense & 96.38$\%$ & 46.53$\%$ \\
        (MinP-Max)+(MaxP-Min)+Flatten+Dense & 95.2$\%$ & 52.77$\%$ \\
        (MinP-S)+Conv+Flatten+Dense & 95.85$\%$ & 55.73$\%$ \\
    \bottomrule
\end{tabular}
}
\vspace{.1in}
\end{table}

From the results of Table~\ref{tab:classification}, we conclude that the networks containing the tropical convolutional structure have much stronger information extraction and expression capabilities than those containing only the ordinary convolution. In particular, on color images, the network that uses both tropical convolution and ordinary convolution achieves the highest accuracy, which also reminds us that the combination of tropical convolution and ordinary convolution on a specific data set can exert their respective advantages.

\subsection{Noise Resistance}\label{4.3}

The classification model's resistance to image noise is also an important part of the model's capabilities. Therefore, we have carried out several noise processing strategies on the CIFAR10 color image data set, namely:

Noise1: Noise with standard normal distribution $*0.01$ added to all points.

Noise2: Noise with standard normal distribution $*0.1$ added to all points.

Noise3: Add noise to a random $5\%$ complete block of the picture.

Noise4: Add noise to random $5\%$ of pixels in the picture.

\begin{table}[H]
    \caption{Changes in accuracy of TCNNs under different noise strategies on CIFAR10}
    \label{tab:robustness}
    \centerline{
    \begin{tabular}{ccccl}
        \toprule
        Network structure         & Noise1 & Noise2 & Noise3 & Noise4 \\
        \midrule
        Conv+Conv+Flatten+Dense   & -0.61$\%$ & -10.17$\%$ & -15.02$\%$ & -10.07$\%$ \\
        (MinP-S)+(MaxP-S)+Flatten+Dense & -0.21$\%$ & -27.79$\%$ & -8.30$\%$  & -10.23$\%$ \\
        (MinP-Max)+(MaxP-Min)+Flatten+Dense     & -0.87$\%$ & -23.27$\%$ & -10.10$\%$ & -8.19$\%$ \\
        (MinP-S)+Conv+Flatten+Dense  & -0.50$\%$ & -28.17$\%$ & -9.74$\%$  & -6.80$\%$ \\
    \bottomrule
\end{tabular}
}
\vspace{.1in}
\end{table}
These noise strategies have been applied to all the test-set images and tested again using pre-trained models of various networks. From the results, we observe that no model can achieve the best robustness on all noise strategies, while those networks adopting tropical convolution modules have better test results than those using only ordinary convolution on Noise3 and Noise4 strategies.

\section{Conclusion}\label{5}
In this paper, we make a new attempt to remove multiplication in the image convolution operation commonly used in computer vision. We've got inspirations from tropical mathematics and min-plus convolution, and proposed TCNNs based on tropical convolution structures for image convolution, including several different similarity measurement operations between images and convolution kernels, and the basic blocks of tropical convolution layers. Moreover, we conduct experiments with real data on these TCNNs, in which we analyze the feature map extracted by the hidden layers in the structure, and evaluate the actual performance of TCNNs on image classification and its robustness to different noises. From the results, compared to the traditional CNNs, TCNNs have stronger non-linear expression ability in general, and its resistance to different noises has its own advantages and disadvantages. From a practical point of view, this proposal of TCNNs provides richer possibilities for us to build deep neural networks. In the future, we can select suitable tropical convolution modules for actual tasks and build networks for specific needs. Regarding future work, we hope to apply the tropical convolution modules to more types of deep neural networks to conduct a more in-depth analysis of this emerging convolution design.

\bibliographystyle{unsrt}
\bibliography{citation}

\end{document}